\documentclass{article}

 \usepackage[preprint]{neurips_2025}

\usepackage[utf8]{inputenc} 
\usepackage[T1]{fontenc}    
\usepackage{hyperref}       
\usepackage{url}            
\usepackage{booktabs}       
\usepackage{amsfonts}       
\usepackage{nicefrac}       
\usepackage{microtype}      
\usepackage{xcolor}         

\usepackage{times}
\usepackage{latexsym}
\usepackage{algorithm}
\usepackage{algorithmic}
\usepackage{pifont}
\usepackage{multirow}
\usepackage{subcaption}
\usepackage{graphicx}
\usepackage{hyperref}

\title{Training Emergent Joint Associations: A Reinforcement Learning Approach to Creative Thinking in Language Models}

\author{
  Mukul Singh \\
  Microsoft, USA \\  \texttt{singhmukul@microsoft.com} \\
  \And
  Ananya Singha \\
  Microsoft, India \\
  \texttt{ananyasingha@microsoft.com} \\
  \And
  Aishni Parab \\
  University of California, Los Angeles, USA \\
  \texttt{   
aishni@g.ucla.edu } \\
  \And
 Pronita Mehrotra \\
  MindAntix \\
  \texttt{pronita@mindantix.com} \\
  \And
  Sumit Gulwani \\
  Microsoft, USA \\
  \texttt{sumitg@microsoft.com} \\
  }

\begin{document}

\maketitle
\begin{abstract}
Associative thinking—the ability to connect seemingly unrelated ideas—is a foundational element of human creativity and problem-solving. This paper explores whether reinforcement learning (RL) guided by associative thinking principles can enhance a model's performance across diverse generative tasks, including story writing, code generation, and chart creation. We introduce a reinforcement learning framework that uses a prompt-based evaluation mechanism, incorporating established divergent thinking metrics from creativity research. A base language model is fine-tuned using this framework to reward outputs demonstrating higher novelty through higher degrees of conceptual connectivity. Interestingly, the experimental results suggest that RL-based associative thinking-trained models not only generate more original and coherent stories but also exhibit improved abstraction and flexibility in tasks such as programming and data visualization. Our findings provide initial evidence that modeling cognitive creativity principles through reinforcement learning can yield more adaptive and generative AI.
\end{abstract}

\section{Introduction}

Large language models (LLMs) are becoming omnipresent in modern AI systems, powering tools for writing, coding, education, design, and decision-making. Their widespread adoption underscores both their impressive generative capacity and their limitations \cite{deepseekai2025deepseekr1incentivizingreasoningcapability, openai2024openaio1card}. Despite their scale and fluency, LLMs often struggle to go beyond pattern reproduction, especially when tasks demand creativity, abstraction, or original synthesis.
Despite their success in domains like code \cite{le2022coderl, li2022alphacode}, math \cite{ahn2024largelanguagemodelsmathematical} and reasoning \cite{reasoning-as-planning}, studies have shown that these models struggle with creative tasks \cite{cog-sci-dont-reason-too-much, cascella2023evaluating}, producing artifacts that show higher semantic similarity compared to humans \cite{dell2025cybernetic, vinchon2024genai}.
The limitation can be explained in the way these models work, where they parameterize the output distribution of tokens conditioned on the sequence of the previous tokens. 
The model learns to greedily optimize the next token, and thus struggles to associate concepts that are spatially distant in the attention maps. 
Their training objectives prioritize the prediction of the next token over the generation of novel ideas, often favoring coherence over surprise \cite{vennam2024rethinkingthinkingtokensunderstanding}.
This is especially more pronounced when the model has not seen those associations before \cite{KAUFMAN2009374}.

Associative thinking—a cognitive process that links seemingly unrelated ideas—is widely recognized as a core mechanism underlying creativity \cite{BEATY2023671, KAUFMAN2009374}. In one of the earliest associative theories of creativity, Sarnoff Mednick \cite{Mednick1962a} proposed a hierarchical model of associativity in which people with steep hierarchies generate less creative responses, while highly creative individuals exhibit flatter hierarchies that enable them to traverse broader semantic distances. The associative process has also been described through an exploration-exploitation framework\cite{BEATY2023671}. Individuals switch between exploratory associative searches (finding new categories) and exploitative associative searches (finding elements within a category).
Associative thinking has been shown to trigger areas of the brain related to episodic and semantic memory \cite{BEATY2023671}. 
It enables humans to generate novel insights, metaphors, and solutions by drawing connections across various domains.

Creativity is a desirable trait for any intelligent system. However, current LLMs are not explicitly optimized for creativity. As a result, they do not exhibit the kind of fluid and flexible thinking associated with human creativity.
Previous research in psychology and neuroscience has shown that associative thinking is measurable and trainable \cite{wu2020systematic}, offering a compelling blueprint for developing more creative AI.

In this work, we explore whether reinforcement learning (RL) can be used to instill associative thinking in LLM, thus enhancing their creative capabilities. We design a novel reward function derived from four dimensions of divergent thinking identified in cognitive science literature \cite{BEATY2023671}, and use it to train multiple base language models through RL. The reward signals encourage the model to produce outputs that reflect deeper, more diverse associations, rather than surface-level coherence. We experiment with models of different sizes and varying base levels of creativity \cite{fein2025litbenchbenchmarkdatasetreliable}.

Our experiments show that models trained with this method outperform baselines in tasks requiring creativity, including story writing, code generation, and chart creation. Qualitative analysis further reveals that RL-enhanced models generate responses that are more imaginative, diverse, and structurally inventive, attributes closely tied to associative thinking.


In summary, we make the following contributions:
\begin{enumerate}
    \item We introduce a reinforcement learning framework that explicitly promotes associative thinking, enabling language models to generate diverse and conceptually distant ideas.
    \item We propose a creativity reward function grounded in cognitive science, and show that it is both robust and strongly aligned with human judgments.
    \item We demonstrate that fostering associative thinking benefits not only traditionally creative domains such as storytelling, but analytical tasks including coding and data visualization.
\end{enumerate}

\section{Related Works}

Recent progress in language models has demonstrated the power of explicit reasoning—via techniques like chain-of-thought prompting, scratchpads, and intermediate program synthesis—to solve complex tasks that require multi-step inference \cite{openai2024openaio1card}. These methods generate intermediate "reasoning tokens" that serve both as a decomposition of the problem and as a form of internal scaffolding. While reasoning has been shown to improve accuracy and interpretability across domains like math \cite{ahn2024largelanguagemodelsmathematical}, code \cite{le2022coderl}, and logic \cite{reasoning-as-planning}, most prior work has focused on prompting or output analysis. In contrast, little attention has been paid to how reasoning behavior evolves during training, or how it relates to internal representations and skill acquisition.

Additionally, several works have been proposed to constrain the behavior of LLMs to produce more precise systematic outputs \cite{wei2022chain, nye2021scratchpad, yao2023tree}. These studies focus on convergent thinking, where the goal is to find a single optimal solution. In contrast, creativity involves divergent thinking, where a problem can have multiple solutions.

Creativity, described as the production of useful and original artifacts, has been studied from various perspectives. The 4P model \cite{Rhodes1961a} identifies four different dimensions from which creativity can be studied: Person (personality traits and abilities that affect creative capacity), Process (mental and behavioral stages involved in producing creative work), Press (social and cultural environment that fosters or stifles creativity), and Product (creativity of the artifact produced). 

Evaluations of creativity from a product perspective include these two primary metrics of originality and usefulness \cite{sternberg1999a, Runco2012a, Diedrich2015a} and sometimes include a third factor of style \cite{Besemer1998a} or wholeness \cite{Henriksen2015a}. Assessing creativity typically involves human subject-domain experts, making it less conducive for automated approaches. 

An alternative approach, of using Guilford metrics \cite{plucker2010assessment}, originally used in the context of Person or Process, has recently been adapted to evaluate AI-generated creative content. The metrics use four core dimensions: fluency (the sheer volume of meaningful ideas produced), flexibility (the diversity of categories within the responses), originality (the uniqueness of answers) and elaboration (the depth or granularity of details within the responses) to estimate overall creativity. 

Several studies have directly applied Guilford's creativity framework to assess AI and LLM capabilities across diverse domains. Stevenson et al. \cite{stevenson2022putting} evaluated GPT-3's creativity on the Alternative Uses Test, comparing the model's performance to human psychology students. Recent applications have extended Guilford's metrics beyond traditional domains. DeLorenzo et al. adapted the four cognitive dimensions to evaluate LLMs within hardware code generation contexts, demonstrating the framework's versatility across technical applications \cite{delorenzo2024creativeval}. Similarly, Elgarf et al. used the four creativity measures to fine-tune GPT-3 models for collaborative storytelling with children \cite{elgarf2024fostering}. In this paper, we apply Guilford's metrics to assess overall creativity, in the reward function of reinforcement learning.

Unlike previous studies, in this paper, we design a novel reward model to measure associative thinking in a model rollout and use this to train language models, showing that such associative training, makes the models better at both conventionally creative (like storytelling \cite{Srivastava2023a}) and non-creative tasks (code generation \cite{cassano2022multipl-e}).

\section{Associative Thinking and RL}

\subsection{Measuring Creativity}

To evaluate the creativity of LLM responses, we treated each association as a “mini idea”. For example, a creative story doesn’t have just one element that makes it creative (although it might) but often employs multiple associations through metaphors or surprising juxtapositions of unexpected concepts. Each of these associations, if they are relevant, contribute to the overall creativity of the story. To estimate the overall creativity of the response, we then applied Guilford’s divergent thinking metrics as described below. 
\vspace{-11pt}
\paragraph{Novelty}: Measures how unusual or unexpected the associations are relative to typical patterns. Higher novelty indicates the model connects concepts rarely seen together.
\vspace{-11pt}
\paragraph{Fluency}: Quantifies the number of distinct associations produced in response to a prompt. Greater fluency reflects the model’s ability to generate many ideas.
\vspace{-11pt}
\paragraph{Flexibility}: Captures the diversity of categories or semantic domains represented in the associations. This dimension evaluates whether the model can traverse multiple conceptual spaces rather than remaining within a narrow topic.
\vspace{-11pt}
\paragraph{Elaboration}: Assesses the level of detail and depth in explaining or expanding on the associations. Rich elaboration signals more developed and meaningful connections.

These dimensions together provide a holistic measure of associative thinking, moving beyond simple lexical diversity or coherence to capture overall creativity.

\subsection{Reinforcement Learning Training}
We fine-tune a base pretrained language model using reinforcement learning (RL) to enhance associative thinking as defined by the above criteria. The RL training employs a policy-gradient algorithm, where the policy is the LLM generating text, and the reward function is derived from the creativity measures.

At each training iteration, the model generates candidate outputs in response to designed prompts targeting associative thinking. These outputs are scored by the reward function, which aggregates the novelty, fluency, flexibility, and elaboration metrics into a scalar reward signal. The model parameters are updated to maximize expected reward, encouraging outputs that demonstrate richer and more diverse associations.

Convergence is monitored via reward stabilization and qualitative improvements in creativity metrics on validation prompts. Early stopping is applied to prevent overfitting and degradation of language fluency.

\subsection{Reward Function}
To implement the reward function, we construct automated graders leveraging the capabilities of large language models. Each grader is designed as a checklist-guided evaluator that inspects candidate outputs along the creativity dimensions.
For example, to measure novelty, the grader prompts the LLM to identify unusual or rare concept combinations within the text. For fluency, it counts distinct associations listed by the model. Flexibility is evaluated by categorizing associations into semantic clusters and scoring their diversity. Elaboration is assessed by requesting detailed explanations or expansions on connections.

We get human participants to annotate 50 samples on these criterias and adjust them for annotater bias.
The checklists were carefully crafted and iteratively refined to align with human judgments, ensuring reliable and interpretable reward signals. This approach enables scalable, automated evaluation of associative thinking without requiring extensive human annotation.

\begin{table}[h]
\small
\centering
\caption{Summary statistics for the benchmarks. We report number of tasks, metrics and the source.}
\begin{tabular}{cccc}
\toprule
\textbf{Task} & \textbf{\#Tasks} & \textbf{Metric Used} & \textbf{Benchmark} \\
\midrule
Storytelling & 5,000 & LLM-based evaluator & \textit{LitBench} \cite{fein2025litbenchbenchmarkdatasetreliable} \\
\addlinespace[2pt]
Code Gen. & 18,000+ & Test pass rate & \textit{MultiPL-E} \cite{cassano2022multipl-e} \\
\addlinespace[2pt]
Visualization & 500 & LLM-based evaluator & Custom (ChartEval) \\
\bottomrule
\end{tabular}
\vspace{-11pt}
\label{tab:benchmark_overview}
\end{table}

\section{Experiment Setup}

\subsection{Training Harness}
We conduct experiments to support the learning hypotheses introduced in the paper. We use reinforcement learning as the primary training method since it is closest to task based knowledge acquisition. We use standard reinforcement learning strategies including (1) Proximal Policy Optimization (PPO) \cite{schulman2017proximalpolicyoptimizationalgorithms}; (2) Group Relative Policy Optimization (GRPO) \cite{deepseekai2025deepseekr1incentivizingreasoningcapability}; (3) Reinforce \cite{mroueh2025reinforcementlearningverifiablerewards}. We train for 100 iterations with 32 rollouts per iteration. We train on a cluster of 8xH100 and the overall training takes 2 hours 35 minutes on average per model.

\subsection{Benchmarks}
We use multiple benchmarks to evaluate both conventionally creative and analytical tasks.
\vspace{-11pt}
\begin{enumerate}
    \item \textbf{Storytelling}: Generating coherent stories from a premise is a creative task even used in high school writing curriculum. We benchmark against the LitBench \cite{fein2025litbenchbenchmarkdatasetreliable} benchmark.
    \item \textbf{Code Generation}: Coding is an inherently analytical task involving composition of mathematical and language constructs. We benchmark against the MultiPL-E \cite{cassano2022multipl-e} benchmark which contains code generation tasks for over 30 high and low-resource languages.
    \item \textbf{Data Visualization}: Visualization and data shaping require a mixture of analytical (data pivots, filter, transformations, etc.) and creative (plot choice, colors, elements, etc.) reasoning.
\end{enumerate}

\subsection{Metrics}
We use different metrics per benchmark to report performance. In particular, (1) for LitBench, we use the proposed LLM based evaluator that was released with the benchmark; (2) for MultiPL-E, we use the test cases that are part of the test-split of the dataset \cite{cassano2022multipl-e}; (3) for charting, we use the chart quality measurement as proposed by \cite{chakrabarty2023art}.

\subsection{Models and Configuration}
We report results on both small and large language models --Deepseek-distill-7B, Phi-4-13B, GPT-4o-mini, GPT-o4-mini. We use the default setup of the model with two variations in inference -- (1) with reasoning which is the standard mode; and (2) without reasoning where the reasoning tokens are turned off and the model behaves like a regular completion model.

\section{Results}
We answer the following research questions:
\begin{itemize}
    \item[\textbf{RQ1.}] Does reinforcing associative thinking improve creativity in language models?
    \item[\textbf{RQ2.}] Does associative thinking reward capture creativity in generations?
    \item[\textbf{RQ3.}] Do gains in associative thinking transfer across non-creative and analytical domains?
\end{itemize}

\subsection{RQ1: Creativity Improvements with RL}
We evaluate the improvement in performance on the benchmarks through associative reward training. For this experiment, we evaluate the models on the benchmark in the base setting and compare it with the trained version. Table~\ref{tab:creativity-results} shows the results across different models.

We can see that the training improves performance across all models and domains (8-13\%) indicating that associativity helps the model in problem solving. Further, the improvement over the storytelling benchmark is the highest which is not surprising since its a conventionally creative task. Code generation being a conventionally analytical task, shows the least improvement but still shows improvement for all models except Phi-3.5-instruct for which it regresses.

\begin{figure}[h]
    \centering
    \begin{subfigure}[t]{0.48\textwidth}
        \centering
        \includegraphics[width=\textwidth]{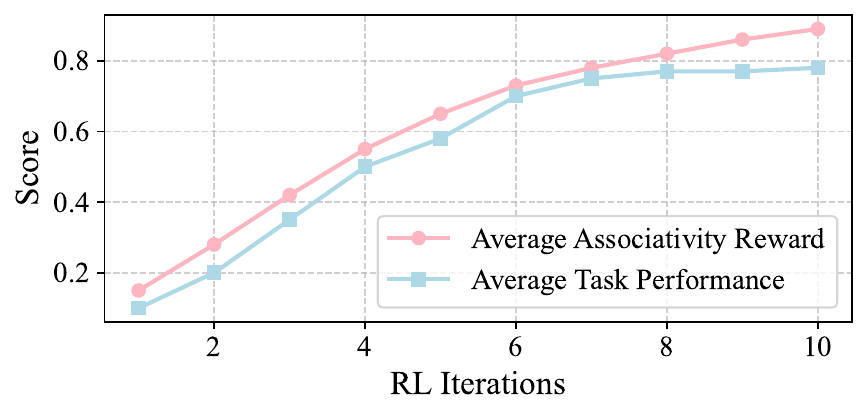}
        \caption{Creativity RL convergence showing alignment between peak associativity reward and task performance.}
        \label{fig:creativity-convergence}
    \end{subfigure}
    \hfill
    \begin{subfigure}[t]{0.48\textwidth}
        \centering
        \includegraphics[width=\textwidth]{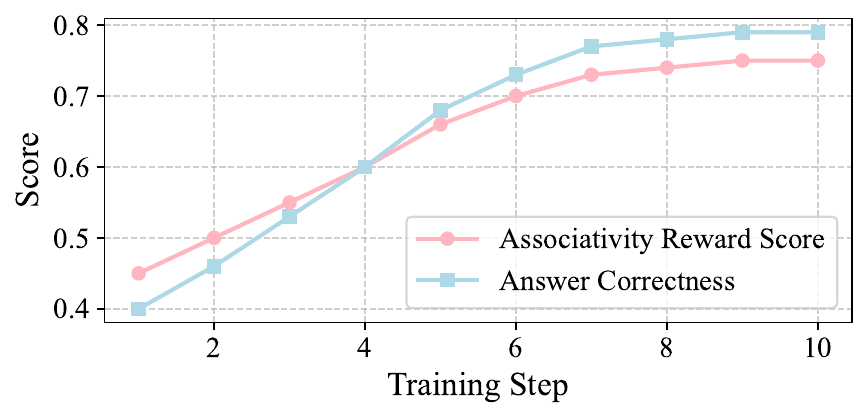}
        \caption{Stability of creativity reward scores and their alignment with task accuracy over training.}
        \label{fig:reasoning-scaffolding}
    \end{subfigure}
    \caption{Analysis of training for associative thinking RL models. (a) Convergence of creativity rewards and task performance. (b) Stability of creativity reward and relationship to task accuracy.}
    \label{fig:training-dynamics}
\end{figure}

We also evaluate the convergence of the training process across RL iterations. Figure~\ref{fig:creativity-convergence} shows the performance of the models against increasing RL iterations. We see that the model rewards and task performance peak around the same region indicating that the model performance gains are aligned to the average associativity reward for its generations.

\begin{table}[h]
\vspace{-11pt}
\centering
\caption{Performance Improvements from Associative Thinking RL Training (\%)}
\label{tab:creativity-results}
\begin{tabular}{lccc}
\toprule
\textbf{Model} & \textbf{Storytelling} & \textbf{Code Generation} & \textbf{Data Visualization} \\
\midrule
Deepseek-distill-7B & 12.1 & 8.3 & 9.5 \\
Phi-4-13B & 13.4 & 7.2 & 10.2 \\
Phi-3.5-instruct & 9.8 & \textbf{-1.5} & 8.7 \\
\bottomrule
\end{tabular}
\end{table}

\subsection{RQ2: Creativity Reward Model}
Since we use a non-deterministic LLM-based reward model for evaluating associativity, it is important to assess the robustness and reliability of this reward scoring mechanism. An effective reward function should be (1) consistent across multiple runs, (2) correlated with human judgments of creativity, and (3) predictive of downstream task performance.
To evaluate this, we conducted repeated evaluations on a fixed set of generations and measured the variance in reward scores. Figure~\ref{fig:reasoning-scaffolding} shows the distribution of reasoning scores and answer correctness as training progresses. The reward scores exhibit low variance, indicating stability despite the underlying stochasticity of the LLM evaluator.
Furthermore, we performed correlation analysis between the associative reward and human-rated creativity scores on a sample of story generations. The results showed a strong positive correlation (Pearson’s $r=0.78$), confirming that the reward captures meaningful creative attributes.

Finally, we observe that improvements in associative reward closely track task accuracy, suggesting that the reward effectively guides the model towards more creative yet relevant generations.

\begin{table}[h]
\centering
\caption{Alignment Between Human Creativity Ratings and Associative Reward Scores}
\label{tab:human-reward-alignment}
\begin{tabular}{lccc}
\toprule
\textbf{Model} & \textbf{Correlation (Pearson's $r$)} & \textbf{Mean Reward Score} & \textbf{Mean Human Rating} \\
\midrule
Deepseek-distill-7B & 0.75 & 0.68 & 3.5 \\
Phi-4-13B           & 0.78 & 0.72 & 3.8 \\
Phi-3.5-instruct         & 0.80 & 0.70 & 3.7 \\
\bottomrule
\end{tabular}
\end{table}

\subsection{RQ3: Transfer to Non-Creative Domains}
To understand if the benefits of associative thinking extend beyond conventionally creative tasks, we evaluate model performance on analytical tasks such as code generation and data visualization.
Table~\ref{tab:transfer-results} presents results comparing base models with associative reward-trained models on these tasks. We observe consistent improvements on data visualization tasks (up to 10\%) indicating that associative thinking aids in creative decision-making like chart design.
However, for code generation, results are more mixed. While most models show modest gains, some models (e.g., Phi-3.5-instruct) experience slight performance degradation, suggesting that enforcing associative creativity might occasionally conflict with the precision required for programming tasks.
These findings highlight that associative thinking can enhance model flexibility and generalization, but careful balancing is necessary when applying it to highly structured, correctness-critical domains.

\begin{table}[h]
\centering
\caption{Transfer of Associative Thinking Gains to Analytical Domains (\%)}
\label{tab:transfer-results}
\begin{tabular}{lcc}
\toprule
\textbf{Model} & \textbf{Code Generation} & \textbf{Data Visualization} \\
\midrule
Deepseek-distill-7B & 7.9 & 9.5 \\
Phi-4-13B & 6.3 & 10.0 \\
Phi-3.5-instruct & -1.2 & 8.5 \\
\bottomrule
\end{tabular}
\vspace{-11pt}
\end{table}

\section{Conclusion}
We demonstrated that training guided by associative thinking principles enhances the creativity of models across multiple domains. Our approach improves performance on storytelling, code generation, and data visualization tasks, with strong alignment between the learned reward and human creativity judgments. While benefits extend to analytical tasks, balancing creativity with accuracy remains important. We show the value of combining cognitive insights with RL for AI creativity.

\section{Limitations}

\paragraph{LLM based creativity graders}
The design of the creativity reward function relies on prompt-based evaluations and checklist-style graders, which, although automated and scalable, may introduce biases or overlook subtleties in human creativity. These graders are based on language models themselves, raising questions about circularity and alignment drift, especially when both training and evaluation depend on similar LLM architectures.

\paragraph{Limited domains in evaluation}
The benchmark coverage, though diverse, is limited to three domains—storytelling, code generation, and data visualization. These were chosen to span a range of cognitive demands, but they do not encompass other important areas where creativity is critical, such as scientific hypothesis generation, musical composition, or social interaction design.

\paragraph{Reinforcement only training}
The reinforcement learning process introduces challenges around stability, sample efficiency, and unintended side effects. Despite reward convergence, we observed occasional degradation in fluency or factual grounding, particularly in code-related outputs. This highlights the need for more robust training strategies that can balance creativity with correctness.


\bibliographystyle{abbrv}

\bibliography{custom}

\begin{thebibliography}{10}

\bibitem{ahn2024largelanguagemodelsmathematical}
J.~Ahn, R.~Verma, R.~Lou, D.~Liu, R.~Zhang, and W.~Yin.
\newblock Large language models for mathematical reasoning: Progresses and challenges, 2024.

\bibitem{BEATY2023671}
R.~E. Beaty and Y.~N. Kenett.
\newblock Associative thinking at the core of creativity.
\newblock {\em Trends in Cognitive Sciences}, 27(7):671--683, 2023.

\bibitem{Besemer1998a}
S.~P. Besemer.
\newblock Creative product analysis matrix: testing the model structure and a comparison among products--three novel chairs.
\newblock {\em Creativity Research Journal}, 11(4):333--346, 1998.

\bibitem{cascella2023evaluating}
M.~Cascella, J.~Montomoli, V.~Bellini, and E.~Bignami.
\newblock Evaluating the feasibility of chatgpt in healthcare: an analysis of multiple clinical and research scenarios.
\newblock {\em Journal of medical systems}, 47(1):33, 2023.

\bibitem{cassano2022multipl-e}
F.~Cassano, J.~Gouwar, D.~Nguyen, S.~Nguyen, L.~Phipps-Costin, D.~Pinckney, M.-H. Yee, Y.~Zi, C.~J. Anderson, M.~Q. Feldman, et~al.
\newblock Multipl-e: a scalable and polyglot approach to benchmarking neural code generation.
\newblock {\em IEEE Transactions on Software Engineering}, 49(7):3675--3691, 2023.

\bibitem{chakrabarty2023art}
T.~Chakrabarty, P.~Laban, D.~Agarwal, S.~Muresan, and C.-S. Wu.
\newblock Art or artifice? large language models and the false promise of creativity.
\newblock {\em arXiv preprint arXiv:2309.14556}, 2023.

\bibitem{deepseekai2025deepseekr1incentivizingreasoningcapability}
DeepSeek-AI, D.~Guo, D.~Yang, H.~Zhang, J.~Song, R.~Zhang, R.~Xu, Q.~Zhu, S.~Ma, P.~Wang, X.~Bi, X.~Zhang, X.~Yu, Y.~Wu, Z.~F. Wu, Z.~Gou, Z.~Shao, Z.~Li, Z.~Gao, A.~Liu, B.~Xue, B.~Wang, B.~Wu, B.~Feng, C.~Lu, C.~Zhao, C.~Deng, C.~Zhang, C.~Ruan, D.~Dai, D.~Chen, D.~Ji, E.~Li, F.~Lin, F.~Dai, F.~Luo, G.~Hao, G.~Chen, G.~Li, H.~Zhang, H.~Bao, H.~Xu, H.~Wang, H.~Ding, H.~Xin, H.~Gao, H.~Qu, H.~Li, J.~Guo, J.~Li, J.~Wang, J.~Chen, J.~Yuan, J.~Qiu, J.~Li, J.~L. Cai, J.~Ni, J.~Liang, J.~Chen, K.~Dong, K.~Hu, K.~Gao, K.~Guan, K.~Huang, K.~Yu, L.~Wang, L.~Zhang, L.~Zhao, L.~Wang, L.~Zhang, L.~Xu, L.~Xia, M.~Zhang, M.~Zhang, M.~Tang, M.~Li, M.~Wang, M.~Li, N.~Tian, P.~Huang, P.~Zhang, Q.~Wang, Q.~Chen, Q.~Du, R.~Ge, R.~Zhang, R.~Pan, R.~Wang, R.~J. Chen, R.~L. Jin, R.~Chen, S.~Lu, S.~Zhou, S.~Chen, S.~Ye, S.~Wang, S.~Yu, S.~Zhou, S.~Pan, S.~S. Li, S.~Zhou, S.~Wu, S.~Ye, T.~Yun, T.~Pei, T.~Sun, T.~Wang, W.~Zeng, W.~Zhao, W.~Liu, W.~Liang, W.~Gao, W.~Yu, W.~Zhang, W.~L. Xiao, W.~An, X.~Liu, X.~Wang, X.~Chen, X.~Nie,
  X.~Cheng, X.~Liu, X.~Xie, X.~Liu, X.~Yang, X.~Li, X.~Su, X.~Lin, X.~Q. Li, X.~Jin, X.~Shen, X.~Chen, X.~Sun, X.~Wang, X.~Song, X.~Zhou, X.~Wang, X.~Shan, Y.~K. Li, Y.~Q. Wang, Y.~X. Wei, Y.~Zhang, Y.~Xu, Y.~Li, Y.~Zhao, Y.~Sun, Y.~Wang, Y.~Yu, Y.~Zhang, Y.~Shi, Y.~Xiong, Y.~He, Y.~Piao, Y.~Wang, Y.~Tan, Y.~Ma, Y.~Liu, Y.~Guo, Y.~Ou, Y.~Wang, Y.~Gong, Y.~Zou, Y.~He, Y.~Xiong, Y.~Luo, Y.~You, Y.~Liu, Y.~Zhou, Y.~X. Zhu, Y.~Xu, Y.~Huang, Y.~Li, Y.~Zheng, Y.~Zhu, Y.~Ma, Y.~Tang, Y.~Zha, Y.~Yan, Z.~Z. Ren, Z.~Ren, Z.~Sha, Z.~Fu, Z.~Xu, Z.~Xie, Z.~Zhang, Z.~Hao, Z.~Ma, Z.~Yan, Z.~Wu, Z.~Gu, Z.~Zhu, Z.~Liu, Z.~Li, Z.~Xie, Z.~Song, Z.~Pan, Z.~Huang, Z.~Xu, Z.~Zhang, and Z.~Zhang.
\newblock Deepseek-r1: Incentivizing reasoning capability in llms via reinforcement learning, 2025.

\bibitem{dell2025cybernetic}
F.~Dell'Acqua, C.~Ayoubi, H.~Lifshitz, R.~Sadun, E.~Mollick, L.~Mollick, Y.~Han, J.~Goldman, H.~Nair, S.~Taub, et~al.
\newblock The cybernetic teammate: A field experiment on generative ai reshaping teamwork and expertise.
\newblock Technical report, National Bureau of Economic Research, 2025.

\bibitem{delorenzo2024creativeval}
M.~DeLorenzo, V.~Gohil, and J.~Rajendran.
\newblock Creativeval: Evaluating creativity of llm-based hardware code generation.
\newblock In {\em 2024 IEEE LLM Aided Design Workshop (LAD)}, pages 1--5. IEEE, 2024.

\bibitem{Diedrich2015a}
J.~Diedrich, M.~Benedek, E.~Jauk, and A.~C. Neubauer.
\newblock Are creative ideas novel and useful?
\newblock {\em Psychology of aesthetics, creativity, and the arts}, 9(1):35, 2015.

\bibitem{elgarf2024fostering}
M.~Elgarf, H.~Salam, and C.~Peters.
\newblock Fostering children’s creativity through llm-driven storytelling with a social robot.
\newblock {\em Frontiers in Robotics and AI}, 11:1457429, 2024.

\bibitem{fein2025litbenchbenchmarkdatasetreliable}
D.~Fein, S.~Russo, V.~Xiang, K.~Jolly, R.~Rafailov, and N.~Haber.
\newblock Litbench: A benchmark and dataset for reliable evaluation of creative writing, 2025.

\bibitem{Henriksen2015a}
D.~Henriksen, P.~Mishra, and R.~Mehta.
\newblock Novel, effective, whole: Toward a new framework for evaluations of creative products.
\newblock {\em Journal of Technology and Teacher Education}, 23(3):455--478, 2015.

\bibitem{KAUFMAN2009374}
S.~B. Kaufman, C.~G. DeYoung, J.~R. Gray, J.~Brown, and N.~Mackintosh.
\newblock Associative learning predicts intelligence above and beyond working memory and processing speed.
\newblock {\em Intelligence}, 37(4):374--382, 2009.

\bibitem{le2022coderl}
H.~Le, Y.~Wang, A.~D. Gotmare, S.~Savarese, and S.~C.~H. Hoi.
\newblock Coderl: Mastering code generation through pretrained models and deep reinforcement learning.
\newblock {\em Advances in Neural Information Processing Systems}, 35:21314--21328, 2022.

\bibitem{li2022alphacode}
Y.~Li, D.~Choi, J.~Chung, N.~Kushman, J.~Schrittwieser, R.~Leblond, T.~Eccles, J.~Keeling, F.~Gimeno, A.~Dal~Lago, et~al.
\newblock Competition-level code generation with alphacode.
\newblock {\em Science}, 378(6624):1092--1097, 2022.

\bibitem{Mednick1962a}
S.~Mednick.
\newblock The associative basis of the creative process.
\newblock {\em Psychological review}, 69(3):220, 1962.

\bibitem{mroueh2025reinforcementlearningverifiablerewards}
Y.~Mroueh.
\newblock Reinforcement learning with verifiable rewards: Grpo's effective loss, dynamics, and success amplification, 2025.

\bibitem{nye2021scratchpad}
M.~Nye, A.~J. Andreassen, G.~Gur-Ari, H.~Michalewski, J.~Austin, D.~Bieber, D.~Dohan, A.~Lewkowycz, M.~Bosma, D.~Luan, et~al.
\newblock Show your work: Scratchpads for intermediate computation with language models.
\newblock {\em arXiv preprint arXiv:2112.00114}, 2021.

\bibitem{openai2024openaio1card}
OpenAI, :, A.~Jaech, A.~Kalai, A.~Lerer, A.~Richardson, A.~El-Kishky, A.~Low, A.~Helyar, A.~Madry, A.~Beutel, A.~Carney, A.~Iftimie, A.~Karpenko, A.~T. Passos, A.~Neitz, A.~Prokofiev, A.~Wei, A.~Tam, A.~Bennett, A.~Kumar, A.~Saraiva, A.~Vallone, A.~Duberstein, A.~Kondrich, A.~Mishchenko, A.~Applebaum, A.~Jiang, A.~Nair, B.~Zoph, B.~Ghorbani, B.~Rossen, B.~Sokolowsky, B.~Barak, B.~McGrew, B.~Minaiev, B.~Hao, B.~Baker, B.~Houghton, B.~McKinzie, B.~Eastman, C.~Lugaresi, C.~Bassin, C.~Hudson, C.~M. Li, C.~de~Bourcy, C.~Voss, C.~Shen, C.~Zhang, C.~Koch, C.~Orsinger, C.~Hesse, C.~Fischer, C.~Chan, D.~Roberts, D.~Kappler, D.~Levy, D.~Selsam, D.~Dohan, D.~Farhi, D.~Mely, D.~Robinson, D.~Tsipras, D.~Li, D.~Oprica, E.~Freeman, E.~Zhang, E.~Wong, E.~Proehl, E.~Cheung, E.~Mitchell, E.~Wallace, E.~Ritter, E.~Mays, F.~Wang, F.~P. Such, F.~Raso, F.~Leoni, F.~Tsimpourlas, F.~Song, F.~von Lohmann, F.~Sulit, G.~Salmon, G.~Parascandolo, G.~Chabot, G.~Zhao, G.~Brockman, G.~Leclerc, H.~Salman, H.~Bao, H.~Sheng, H.~Andrin,
  H.~Bagherinezhad, H.~Ren, H.~Lightman, H.~W. Chung, I.~Kivlichan, I.~O'Connell, I.~Osband, I.~C. Gilaberte, I.~Akkaya, I.~Kostrikov, I.~Sutskever, I.~Kofman, J.~Pachocki, J.~Lennon, J.~Wei, J.~Harb, J.~Twore, J.~Feng, J.~Yu, J.~Weng, J.~Tang, J.~Yu, J.~Q. Candela, J.~Palermo, J.~Parish, J.~Heidecke, J.~Hallman, J.~Rizzo, J.~Gordon, J.~Uesato, J.~Ward, J.~Huizinga, J.~Wang, K.~Chen, K.~Xiao, K.~Singhal, K.~Nguyen, K.~Cobbe, K.~Shi, K.~Wood, K.~Rimbach, K.~Gu-Lemberg, K.~Liu, K.~Lu, K.~Stone, K.~Yu, L.~Ahmad, L.~Yang, L.~Liu, L.~Maksin, L.~Ho, L.~Fedus, L.~Weng, L.~Li, L.~McCallum, L.~Held, L.~Kuhn, L.~Kondraciuk, L.~Kaiser, L.~Metz, M.~Boyd, M.~Trebacz, M.~Joglekar, M.~Chen, M.~Tintor, M.~Meyer, M.~Jones, M.~Kaufer, M.~Schwarzer, M.~Shah, M.~Yatbaz, M.~Y. Guan, M.~Xu, M.~Yan, M.~Glaese, M.~Chen, M.~Lampe, M.~Malek, M.~Wang, M.~Fradin, M.~McClay, M.~Pavlov, M.~Wang, M.~Wang, M.~Murati, M.~Bavarian, M.~Rohaninejad, N.~McAleese, N.~Chowdhury, N.~Chowdhury, N.~Ryder, N.~Tezak, N.~Brown, O.~Nachum, O.~Boiko,
  O.~Murk, O.~Watkins, P.~Chao, P.~Ashbourne, P.~Izmailov, P.~Zhokhov, R.~Dias, R.~Arora, R.~Lin, R.~G. Lopes, R.~Gaon, R.~Miyara, R.~Leike, R.~Hwang, R.~Garg, R.~Brown, R.~James, R.~Shu, R.~Cheu, R.~Greene, S.~Jain, S.~Altman, S.~Toizer, S.~Toyer, S.~Miserendino, S.~Agarwal, S.~Hernandez, S.~Baker, S.~McKinney, S.~Yan, S.~Zhao, S.~Hu, S.~Santurkar, S.~R. Chaudhuri, S.~Zhang, S.~Fu, S.~Papay, S.~Lin, S.~Balaji, S.~Sanjeev, S.~Sidor, T.~Broda, A.~Clark, T.~Wang, T.~Gordon, T.~Sanders, T.~Patwardhan, T.~Sottiaux, T.~Degry, T.~Dimson, T.~Zheng, T.~Garipov, T.~Stasi, T.~Bansal, T.~Creech, T.~Peterson, T.~Eloundou, V.~Qi, V.~Kosaraju, V.~Monaco, V.~Pong, V.~Fomenko, W.~Zheng, W.~Zhou, W.~McCabe, W.~Zaremba, Y.~Dubois, Y.~Lu, Y.~Chen, Y.~Cha, Y.~Bai, Y.~He, Y.~Zhang, Y.~Wang, Z.~Shao, and Z.~Li.
\newblock Openai o1 system card, 2024.

\bibitem{plucker2010assessment}
J.~A. Plucker, M.~C. Makel, and M.~Qian.
\newblock Assessment of creativity.
\newblock {\em The Cambridge handbook of creativity}, pages 48--73, 2010.

\bibitem{Rhodes1961a}
M.~Rhodes.
\newblock An analysis of creativity.
\newblock {\em The Phi delta kappan}, 42(7):305--310, 1961.

\bibitem{Runco2012a}
M.~A. Runco and G.~J. Jaeger.
\newblock The standard definition of creativity.
\newblock {\em Creativity research journal}, 24(1):92--96, 2012.

\bibitem{schulman2017proximalpolicyoptimizationalgorithms}
J.~Schulman, F.~Wolski, P.~Dhariwal, A.~Radford, and O.~Klimov.
\newblock Proximal policy optimization algorithms, 2017.

\bibitem{Srivastava2023a}
S.~Srivastava, S.~Oberoi, and V.~K. Gupta.
\newblock The story and the storyteller: Strategic storytelling that gets human attention for entrepreneurs.
\newblock {\em Business Horizons}, 66(3):347--358, 2023.

\bibitem{sternberg1999a}
R.~J. Sternberg and T.~I. Lubart.
\newblock The concept of creativity: Prospects and paradigms.
\newblock {\em Handbook of creativity}, 1(3-15), 1999.

\bibitem{stevenson2022putting}
C.~Stevenson, I.~Smal, M.~Baas, R.~Grasman, and H.~van~der Maas.
\newblock Putting gpt-3's creativity to the (alternative uses) test.
\newblock {\em arXiv preprint arXiv:2206.08932}, 2022.

\bibitem{vennam2024rethinkingthinkingtokensunderstanding}
S.~Vennam, D.~Valente, D.~Herel, and P.~Kumaraguru.
\newblock Rethinking thinking tokens: Understanding why they underperform in practice, 2024.

\bibitem{vinchon2024genai}
F.~Vinchon, V.~Gironnay, and T.~Lubart.
\newblock Genai creativity in narrative tasks: Exploring new forms of creativity.
\newblock {\em Journal of Intelligence}, 12(12):125, 2024.

\bibitem{reasoning-as-planning}
X.~Wang, L.~Caccia, O.~Ostapenko, X.~Yuan, W.~Y. Wang, and A.~Sordoni.
\newblock Guiding language model reasoning with planning tokens, 2024.

\bibitem{wei2022chain}
J.~Wei, X.~Wang, D.~Schuurmans, M.~Bosma, F.~Xia, E.~Chi, Q.~V. Le, D.~Zhou, et~al.
\newblock Chain-of-thought prompting elicits reasoning in large language models.
\newblock 35:24824--24837, 2022.

\bibitem{cog-sci-dont-reason-too-much}
T.~Wilson and J.~Schooler.
\newblock Thinking too much: Introspection can reduce the quality of preferences and decisions.
\newblock {\em Journal of personality and social psychology}, 60:181--92, 03 1991.

\bibitem{wu2020systematic}
C.-L. Wu, S.-Y. Huang, P.-Z. Chen, and H.-C. Chen.
\newblock A systematic review of creativity-related studies applying the remote associates test from 2000 to 2019.
\newblock {\em Frontiers in psychology}, 11:573432, 2020.

\bibitem{yao2023tree}
S.~Yao, D.~Yu, J.~Zhao, I.~Shafran, T.~L. Griffiths, Y.~Cao, and K.~Narasimhan.
\newblock Tree of thoughts: Deliberate problem solving with large language models.
\newblock {\em arXiv preprint arXiv:2305.10601}, 2023.

\end{thebibliography}



\end{document}